%% file: main.tex
\documentclass[runningheads]{llncs}
\usepackage{graphicx}
\input{config}

\begin{document}
\title{Potential and Limitations of LLMs in Capturing Structured Semantics: A Case Study on SRL}
%
%
\author{Ning Cheng\inst{1,\hspace{0.15cm}\star} \and
Zhaohui Yan\inst{2,} \thanks{These authors contributed equally to this research.} \and
Ziming Wang\inst{1} \and
Zhijie Li\inst{3} \and
Jiaming Yu\inst{4} \and
Zilong Zheng\inst{5} \and
Kewei Tu\inst{2} \and
Jinan Xu\inst{1} \and
Wenjuan Han\inst{1}
}
\authorrunning{Cheng et al.}
%
\institute{Beijing Jiaotong University, Beijing, China \and
ShanghaiTech University, Shanghai, China \and
Tsinghua University, Beijing, China \and
University of Oxford, Oxford, United Kingdom \and
Beijing Institute for General Artificial Intelligence, Beijing, China
}

%
\maketitle              
\begin{abstract}
Large Language Models (LLMs) play a crucial role in capturing structured semantics to enhance language understanding, improve interpretability, and reduce bias. Nevertheless, an ongoing controversy exists over the extent to which LLMs can grasp structured semantics.
To assess this, we propose using Semantic Role Labeling (SRL) as a fundamental task to explore LLMs' ability to extract structured semantics. In our assessment, we employ the prompting approach, which leads to the creation of our few-shot SRL parser, called \model. 
\model enables LLMs to map natural languages to explicit semantic structures, which provides an interpretable window into the properties of LLMs. We find interesting potential: LLMs can indeed capture semantic structures, and scaling-up doesn't always mirror potential. Additionally, limitations of LLMs are observed in C-arguments, \etc. Lastly, we are surprised to discover that significant overlap in the errors is made by both LLMs and untrained humans, accounting for almost 30\% of all errors.

\end{abstract}

\section{Introduction}
\label{sec:intro}

Ongoing innovations in Large language models (LLMs) have provided an influx of apparently state-of-the-art results on various NLP tasks~\cite{brown2020language,ouyang2022training,weichain}. 
Meanwhile, the integration of semantic structures with LLMs brings comprehension of words and phrases beyond simple associations~\cite{shin2021constrained,shin2022few}.
This enhanced capability improves interpretability~\cite{kim2018interpretability}, reduces biases, addresses ethical concerns~\cite{corbett2018measure}, and facilitates effective human-machine collaboration~\cite{alishahi2019analyzing}.

Prior studies~\cite{tenney2019you,he2020establishing,vilares2020parsing} provide insights into structured semantics in LLMs. However, systematic assessment is lacking regarding LLMs' understanding of structured semantics and to what extent they capture the structured semantics of \textbf{open-domain} natural languages \textbf{without} external annotations.

To address this gap, our study comprehensively evaluates state-of-the-art LLMs in the Semantic Role Labeling (SRL) task, aiming to recover structured semantics for a given sentence~\cite{palmer2010semantic}. 
Specifically, we investigate the potential and limitations of LLMs in capturing SRL structures regarding open-domain languages.
Since we perform the assessment using a prompting-based approach, the by-product of this assessment leads to the creation of our few-shot SRL parser (\ie, \model, in Sec. \ref{sec:method}). 
\model performs SRL via inference alone by conditioning on the prompt to demonstrate the inner capability of the LLMs. It's appealing because the prompt design of \model ties to the linguistic theory and matches the predicate-argument structure annotation schemes, similar to QA-SRL~\cite{fitzgerald2018large}, to be intuitive to untrained humans and LLMs. Experiments of LLMs, including Llama2-7B-Chat, ChatGLM2-6B, four variants of GPT-3 and ChatGPT on CoNLL-2005 and CoNLL-2012 (Sec. \ref{sec:exp}), reveal interesting findings: (i) LLMs can indeed capture semantic structures and scaling-up of LLMs doesn't always reflects potential (Sec. \ref{sec:exp_propotional}); (ii) LLMs still struggle with grasping complex semantics (Sec. \ref{sec:exp_can_not_cature_role}); (iii) Trade-off exists between more exemplars and longer prompt (Sec. \ref{sec:exp_exemplar}); (iv) Up to almost 30\% of mistakes are made by both LLMs and untrained humans (Sec. \ref{sec:exp_compr_human_llm}).

\section{Related Work}

\subsection{Few-shot Semantic Role Labeling}
Semantic role labeling~(SRL), a typical semantic parsing task, is to extract predicate-argument structure for natural language with the goal of answering the question: ``when'', ``where'', ``who'', ``what'' and ``to whom'', \etc. 
The disadvantage of supervised SRL is that both the massive annotated training data and highly skilled and trained annotators are required to label the dataset. High cost motivates the need for using unlabeled data or other forms of weak supervision. The few-shot SRL model was first proposed by Swier and Stevenson who use a verb lexicon VerbNet and a supervised syntactic parser \cite{swier-stevenson-2004-unsupervised}. Grenager and Manning \cite{grenager-manning-2006-unsupervised} use EM algorithm for the few-shot learning of a structured probabilistic model. Few-shot semantic role induction problem has been studied in some works
~\cite{titov-klementiev-2012-bayesian,titov-khoddam-2015-unsupervised} and the related task of few-shot argument identification was considered in \cite{abend-etal-2009-unsupervised}.
More recently, Drozdov \etal \cite{drozdov2022compositional} solve semantic parsing tasks by decomposing the problem using prompting-based syntactic parsing and a new state of art for CFQ. However, they still require 1\% of the training data. More importantly, they focus on the area of specialized application instead of an open-domain and daily-life dataset. In our work, we identify these additional challenges in more general semantic parsing tasks without the help of any labeled data. We focus on open-domain natural languages rather than languages in an area of a specialized application.

\subsection{Structured Information in Language Models}
There exists research providing valuable insights into the topic of capturing structured information in NLP models, especially for syntactic information.
Clark \etal \cite{clark2019does} explore the attention mechanism of BERT  and investigates how it captures structured syntax.
Linzen \etal \cite{linzen2016assessing} examine whether LSTM-based language models can capture structured syntax-sensitive dependencies similar to humans. It presents experimental results that compare LSTM models against human performance, shedding light on the models' ability to understand structured syntax.
Additionally, Melis \etal \cite{melisstate} review the evaluation methods used for neural language models and discusses their limitations in capturing structured information. It highlights the challenges of evaluating syntax and semantics and proposes potential directions for improving evaluation practices.

\subsection{Prompt-based Learning}
Recently, prompt-based learning is receiving increasing attention as a general few-shot approach. 
Prompt-based learning reformulates the downstream tasks  with the help of a textual prompt. 
The prompt template could be created manually~\cite{DBLP:conf/nips/BrownMRSKDNSSAA20,schick-schutze-2021-shot,schick-schutze-2021-exploiting,schick-schutze-2021-just} or through an automatic template design process. The automatically induced prompts could be discrete (discrete prompts also known as hard prompts)~\cite{shin-etal-2020-autoprompt,jiang2020can} or continuous~\cite{lester-etal-2021-power,shin-etal-2021-constrained}. Prompt-based learning has been used for semantic parsing. Shin \etal \cite{shin-etal-2021-constrained} explore the use of LMs as few-shot semantic parsers. It tackles semantic parsing as paraphrasing and uses the prompt to make an LM map natural utterances to canonical utterances. Lester \etal \cite{lester-etal-2021-power} empirically investigate prompt tuning for low-resource semantic parsing. Instead of model tuning, it proposed a soft tunable prompt and find that the prompt tuning significantly outperforms the fine-tuned counterpart for LLMs. 

\section{PromptSRL}
\label{sec:method}

\begin{figure*}[htp]
    \small
        \centering       
        \includegraphics[width=1.0\textwidth]{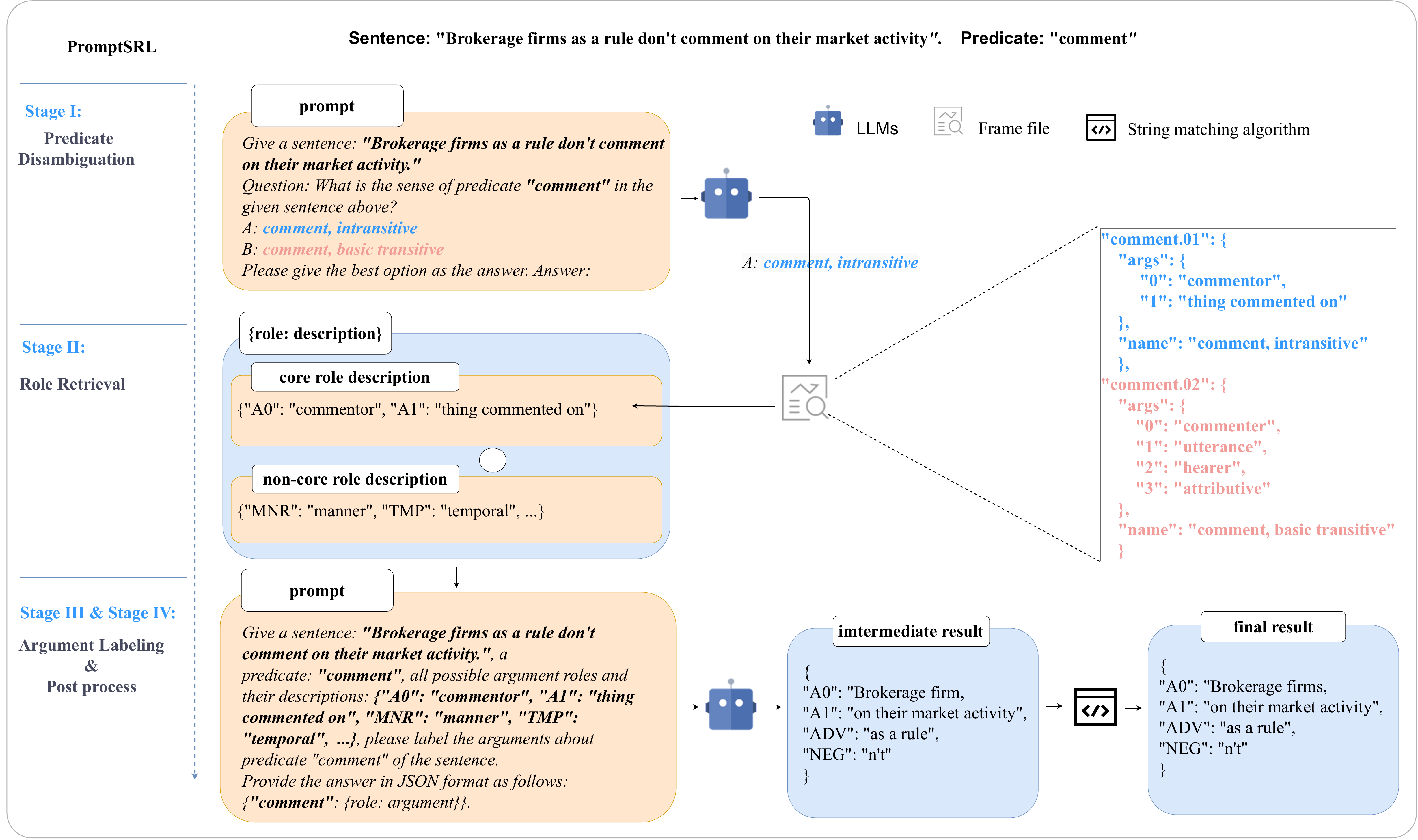}

        \caption{PromptSRL pipeline.}       
        \label{fig:pipeline}
    \end{figure*}

\model performs SRL in a four-stage process: predicate disambiguation, role retrieval, argument labeling, and post-process. The pipeline is shown in Fig. \ref{fig:pipeline}.

\subsection{Stage I: Predicate Disambiguation}
Since the core argument roles \texttt{A0}-\texttt{A5} and \texttt{AA} are predicate-specific, we tie linguistic theory to disambiguate predicates using a multiple-choice task and design a prompt for it to be processed by LLMs. 
Specifically, we provide a sentence with a marked predicate and all possible sense options of the predicate, and ask LLMs to determine the most suitable option. 
Possible sense options are obtained by identifying the lemma of the predicate and then finding all the senses defined under this lemma from the frame files\footnote{\url{https://github.com/ShannonAI/MRC-SRL/tree/main/data}}.
The designed prompt template is as follows: 

\vspace{0.1cm}
\noindent\fbox{
  \parbox{335pt}{
  \fontsize{9.8pt}{12pt}\selectfont
    \emph{Give a sentence: \textless{}sentence\textgreater{}.\\ Question: What is the sense of predicate \textless{}predicate\textgreater{} in the given sentence above? \\ A: \textless{}option 1\textgreater{} \\ B: \textless{}option 2\textgreater{} \\ \dots \\ Please give the best option as the answer. Answer:}
}}

\vspace{0.1cm}
\noindent where \textit{\textless{}option 1\textgreater{}} and \textit{\textless{}option 2\textgreater{}} are the possible senses of \textit{\textless{}predicate\textgreater{}}.


\subsection{Stage II: Role Retrieval}
After Stage I, the sense of the predicate is selected. 
We provide all the available core argument roles and their descriptions associated with the sense of the predicate according to the frame files.
All possible roles and their corresponding descriptions can be acquired for the predicate by uniting the available core roles and all the fixed non-core roles\footnote{Fixed non-core roles can be found in CoNLL-2005 \cite{carreras2005introduction} and CoNLL-2012 \cite{pradhan2012conll}.}.

\subsection{Stage III: Argument Labeling}
We adopt prompt-based inference with LLMs. We use LLMs to identify the arguments of the predicate and assign them their corresponding semantic roles through a designed prompt. Specifically, given a sentence with a marked predicate and a JSON format of possible argument roles taken from Stage II\footnote{\model will ask about all known roles obtained by the role retrieval stage, not all roles from the PropBank framesets.}, we generate a prompt by filling in the frame description of argument roles, rather than using the symbolic abbreviation. The frame description comes from PropBank~\cite{Kingsbury03propbank}, a classic lexical resource for SRL. This makes LLMs perceive semantic information better. The generated prompts are shown in the following format:

\vspace{0.1cm}
\noindent\fbox{
  \parbox{335pt}{
  \fontsize{9.8pt}{12pt}\selectfont
    \emph{Give a sentence: \textless{}sentence\textgreater{}, a predicate: \textless{}predicate\textgreater{}, all possible argument roles and their descriptions: \textless{}\{role: description\}\textgreater{}, please label the arguments about predicate \textless{}predicate\textgreater{} of the sentence. \\
    Provide the answer in JSON format as follows: \{\textless{}predicate\textgreater{}: \{role: argument\}\}. }
}}
\vspace{0.1cm}

\noindent We use the generated prompt for LLMs to complete the answer and extract arguments from the answer.

\subsection{Stage IV: Post-process}
The model sometimes produces a few errors in the prediction, such as \textit{``thomas''} while the gold is \textit{``Thomas''}, though they're pretty close to the correct answer. Therefore, to obtain a precise answer, we try a fuzzy string matching algorithm to refine those predictions and match the most similar phrase components in the given sentence. The string matching algorithm is described in Algorithm~\ref{alg:str_match}.


\renewcommand{\algorithmicrequire}{\textbf{Input:}}
\renewcommand{\algorithmicensure}{\textbf{Output:}}

\begin{algorithm}[!ht]
    \small
    \caption{String matching algorithm}\label{alg:str_match}
    \begin{algorithmic}[1]
        \REQUIRE query string $s_q$, corpus string $s_c$, step value $t$, flexity $f$, SequenceMatcher function $SM$ from difflib which return a similarity value.
        \ENSURE the variable-length substring $s_m$ from $s_c$ that is most similar to $s_q$
        \STATE Initialization $lq \gets $ len($s_q$),  $m\gets 0$, $bl, br\gets 0$, similarity $sim, sim_l, sim_r \gets 0$
        \WHILE{$m + lq -t < $ len($s_c$)}
            \IF{$SM(s_q, s_c[m:m+lq])>sim$}
                \STATE $sim \gets SM(s_q, s_c[m:m+lq])$, \\$bl=m, br=m+lq$, \\$m+=t$
            \ENDIF
        \ENDWHILE
        \STATE $sim_l, sim_r \gets sim$, $tl\gets bl, tr\gets br$
        \FOR{$i=1$ to $f$}
            \IF{$SM(s_q,s_c[tl-i:tr]>sim_l$}
                \STATE $sim_l \gets SM(s_q,s_c[tl-i:tr]$, $bl\gets tl-i$
            \ENDIF
            \IF{$SM(s_q,s_c[tl+i:tr]>sim_l$}
                \STATE $sim_l \gets SM(s_q,s_c[tl+i:tr]$, $bl\gets tl+i$
            \ENDIF
            \IF{$SM(s_q,s_c[tl:tr-i]>sim_r$}
                \STATE $sim_r \gets SM(s_q,s_c[tl:tr-i]$, $br\gets tr-i$
            \ENDIF
            \IF{$SM(s_q,s_c[tl:tr+i]>sim_l$}
                \STATE $sim_r \gets SM(s_q,s_c[tl:tr+i]$, $br\gets tr+i$
            \ENDIF
        \ENDFOR
        \STATE \textbf{return} $s_c[bl:br]$
    \end{algorithmic}
\end{algorithm}


\section{Experiment}
\label{sec:exp}

\subsection{Dataset}
To verify the effectiveness of our aforementioned \model, We conduct experiments on CoNLL-2005~\cite{carreras2005introduction} and CoNLL-2012~\cite{pradhan2012conll} test sets. 

\subsection{Setup}\label{sec.exp_setup}
We adopt Llama2-7B-Chat~\cite{touvron2023llama}, ChatGLM2-6B\footnote{\url{https://github.com/THUDM/ChatGLM2-6B?tab=readme-ov-file}}, ChatGPT\footnote{May 24, 2023 Version; \url{https://platform.openai.com/docs/models/gpt-3-5}} and four variants of GPT-3\footnote{June 2023 Version; \url{https://beta.openai.com/docs/models/gpt-3}}: \textit{text-ada-001}, \textit{text-babbage-001}, \textit{text-curie-001}, \textit{text-davinci-001}. Brown \etal~\cite{brown2020language} called them Ada, Babbage, Curie, and Davinci and implied that their sizes line up closely with 350M, 1.3B, 6.7B, and 175B, respectively. The few-shot setting was only used for the argument labeling stage. 
First, we processed the training dataset into prompt-compatible formats. We then randomly chose few-shot exemplars from the training dataset. Notably, the labeled data are only used as exemplars because \model is finetuning-free.
For the baseline model CoT~\cite{weichain}, we use prompts with symbolic abbreviations and adopt Davinci. We set the temperature to 0 and use micro F1 as metrics. 

\section{Results}\label{sec:result}

\subsection{LLMs' Potential}\label{sec:exp_propotional}
\begin{figure}
    \small
        \centering       
        \includegraphics[width=0.8\linewidth]{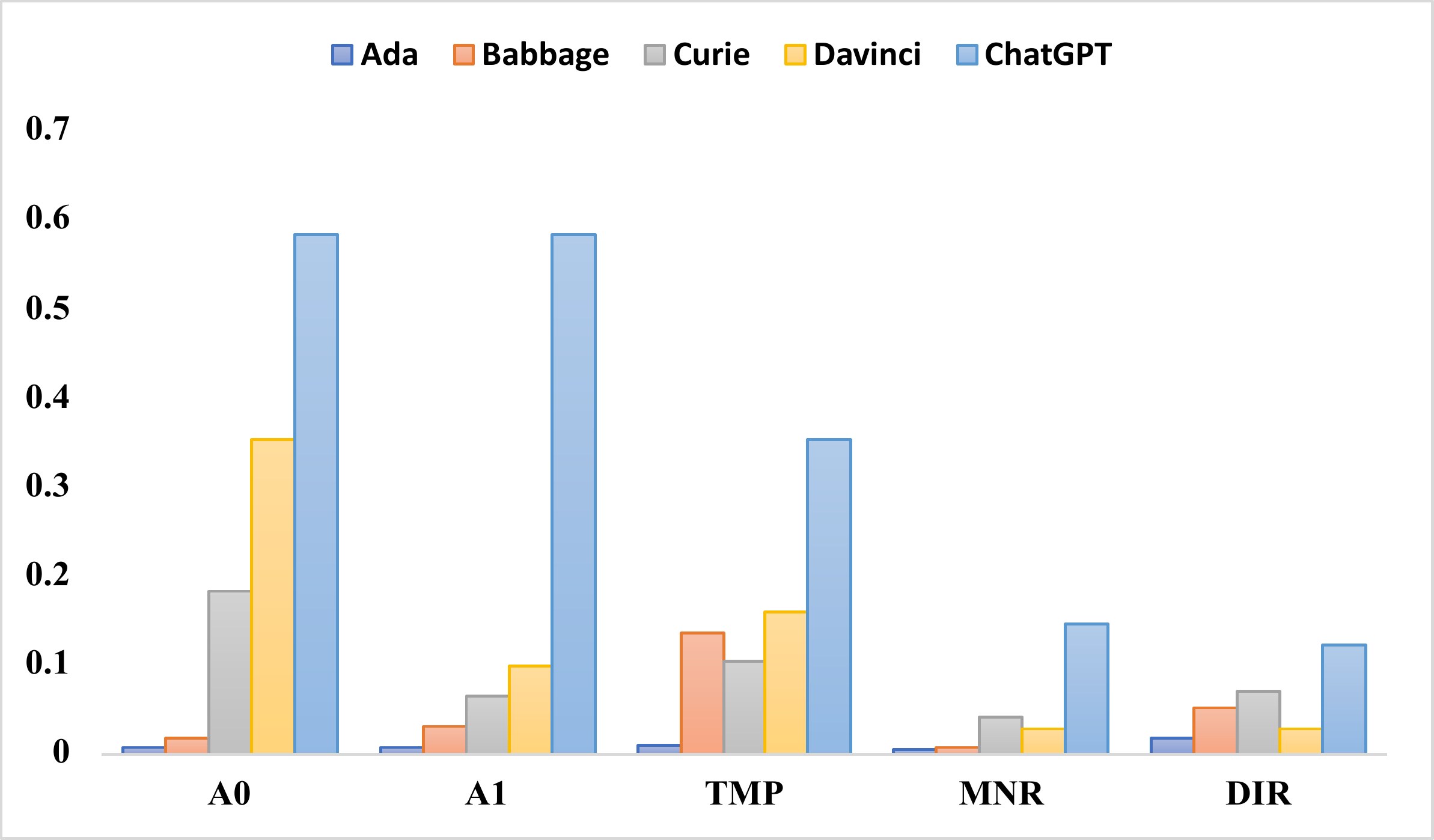}

        \caption{Analysis of selected roles. We use the 3-shot setting and micro F1 for four variants of GPT-3 and ChatGPT.}       
        \label{fig:variants}
    \end{figure}

To systematically and quantitatively validate whether LLMs exhibit semantic understanding ability, we conduct experiments on LLMs with various scales as described in Sec. \ref{sec.exp_setup}. The results are shown in Table \ref{tab:main-result}. In the same few-shot setting, LLMs generally outperform state-of-the-art supervised models. We find that SRL performances are not always proportional to LLMs' scales but reflect the varying capabilities of LLMs in capturing semantics with natural instructions (\ie, prompts). 
Notably, since both CoNLL-2005 and CoNLL-2012 do not have the ground truth of predicate senses, we follow the previous work~\cite{wang2022mrc} and do not evaluate the performance of predicate disambiguation. We then analyze the performance of GPT series, the highly popular LLMs, on each argument role in Fig.~\ref{fig:variants}. Clearly, the scores of most roles increase while scaling up the model.
Additionally, we evaluate the performance of ChatGPT with five different prompts on CoNLL-2005 WSJ in Sec. \ref{app:robust}, providing insights into the robustness of \model. 

\begin{table}[!ht]
\small
\centering
\caption{Comparison of supervised approaches (\ie, HeSyFu~\cite{fei2021better}, CRF2o~\cite{zhang2021semantic}, MRC-SRL~\cite{wang2022mrc}), few-shot approaches (\ie, MRC-SRL~\cite{wang2022mrc} n-shot and CoT) and \model on CoNLL-2005 and CoNLL-2012 test sets.}
\label{tab:main-result}
\begin{tabular}{|c|c|ccc|ccc|ccc|}
\hline
\multirow{2}{*}{\textbf{Model}} & \multirow{2}{*}{\textbf{Shot}} & \multicolumn{3}{c|}{\textbf{CoNLL05 WSJ}} & \multicolumn{3}{c|}{\textbf{CoNLL05 Brown}} & \multicolumn{3}{c|}{\textbf{CoNLL12 Test}} \\ 
\cline{3-11}
 &  & P & R & F1  & P & R & F1 & P & R & F1 \\ 
                     \hline
\makecell[l]{HeSyFu}                               & - & 88.86 & 89.28 & 89.04 & 83.52 & 83.75 & 83.67 & 88.09 & 88.83 & 88.59 \\ 
\makecell[l]{CRF2o}                          & - & 89.45 & 89.63 & 89.54 & 83.89 &  83.39 & 83.64 & 88.11 &  88.53 &  88.32 \\
\makecell[l]{MRC-SRL} 
& - & 90.4 & 89.7 & 90.0 &  86.4 & 83.8 & 85.1 & 88.6 & 87.9 & 88.3 \\
                      \hline
\multirow{6}{*}{MRC-SRL}
& 1-shot & 0.10 & 0.50 & 0.17 & 0.06 & 0.36 & 0.10 & 0.03 & 0.23 & 0.05 \\
& 3-shot &  0.04 & 0.50 & 0.07 & 0.03 & 0.27 & 0.05 & 0.01 & 0.03 & 0.02 \\
& 5-shot & 0.04 & 0.45 & 0.07 & 0.02 & 0.23 & 0.04 & 8.65 & 0.19 & 0.37 \\
& 10-shot & 0.05 & 0.57 & 0.08 & 0.03 & 0.27 & 0.05 & 15.78 & 2.14 & 3.77 \\
& 20-shot & 16.50 & 2.30 & 4.04 & 21.15 & 1.50 & 2.80 & 31.44 & 8.53 & 13.42 \\
& full-shot & 90.34 & 89.58 & 89.96 & 85.47 & 83.80 & 84.62 & 88.52 & 88.39  & 88.45 \\
                    \hline
\makecell[l]{CoT \\ (\textbf{Davinci})} & 3-shot & 6.29 & 26.06 & 10.13 & 4.01 & 18.70 & 6.60 & 2.50 & 16.13 & 4.33 \\
                    \hline
\makecell[l]{\model \\ (\textbf{Llama2-7B-Chat})} & 3-shot & 5.49 & 16.46 & 8.23 & 5.34 & 14.58 & 7.82 & 1.83 & 10.67 & 3.13 \\ 
\makecell[l]{\model \\ (\textbf{ChatGLM2-6B})} & 3-shot & 12.72 & 34.12 & 18.53 & 8.94 & 22.83 & 12.85 & 5.68 & 29.58 & 9.53 \\   
                    \hline
\makecell[l]{\model \\ (\textbf{Ada})} & 3-shot & 2.00 & 1.41 & 1.65 & 0.98 & 0.75 & 0.85 & 0.47 & 0.83 & 0.60 \\                       
\makecell[l]{\model \\ (\textbf{Babbage})} & 3-shot & 3.45 & 1.41 & 2.00 & 3.94 & 3.73 & 3.83 & 4.64 & 7.44 & 5.72 \\                     
\makecell[l]{\model \\ (\textbf{Curie})} & 3-shot & 3.74 & 5.63 & 4.49 & 6.36 & 11.19 & 8.11 & 3.67 & 7.44 & 4.92 \\                          
\makecell[l]{\model \\ (\textbf{Davinci})} & 3-shot & 12.07 & 14.79 & 13.29 & 7.13 & 21.64 & 10.73 & 4.08 & 15.70 & 6.48 \\                          
\makecell[l]{\model \\ (\textbf{ChatGPT})} 
& 3-shot & 39.19 & 41.73 & 40.42 & 37.59 & 41.32 & 39.37 & 36.57  & 40.83 & 38.58 \\
                     \hline
\end{tabular}
\end{table}

\subsection{LLMs' Limitations}\label{sec:exp_can_not_cature_role}

\subsubsection{Discontinuity with C-arguments and R-arguments} According to the definition from CoNLL-2005\footnote{\url{https://www.cs.upc.edu/~srlconll/}}, C-argument describes a continuation phrase of a previously started argument, and R-argument describes a reference to some other argument of \texttt{A*} type (\ie, \texttt{A0}-\texttt{AA}). For example, given the sentence ``\textit{One troubling aspect of DEC's results, analysts said, was its performance in Europe.}'', the \texttt{A1} of \textit{``said''} is split into discontinuous phrases: ``\textit{One troubling aspect of DEC's results}'' and ``\textit{was its performance in Europe}'', so the head phrase is labeled with \texttt{A1} while the non-head one with \texttt{C-A1}. 
    Given the sentence ``\textit{The deregulation of railroads and trucking companies {``that''} began in 1980 enabled shippers to bargain for transportation.}'', {\textit{``that''}} constitutes an argument that actually is referencing another argument, \textit{``the deregulation''}. {\textit{``that''}} is marked as \texttt{R-A1} because \textit{``the deregulation''} is marked as \texttt{A1} of \textit{``began''}. 
    \model meets trouble when dealing with these two uncommon arguments.

\begin{figure}[t]
        \centering
        \includegraphics[width=\linewidth]{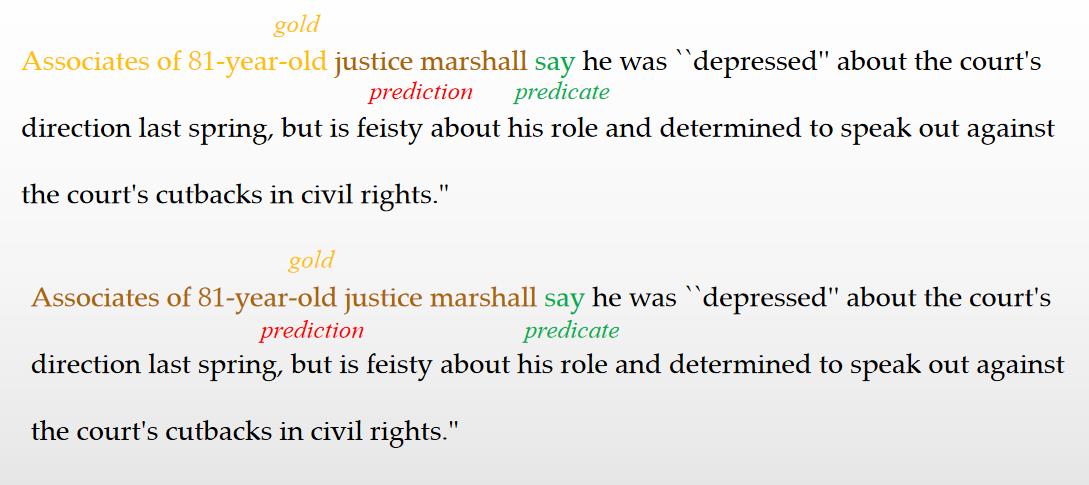}
        \caption{Case study of a successful case. The result at the top is predicted by the baseline model and the result at the bottom is predicted by \model. The baseline model misunderstands the ``Associates of 81-year-old'' part compared to \model.}
        \label{fig:case_study_full_success}
    \end{figure}

\begin{figure}[t]
        \centering
        \includegraphics[width=\linewidth]{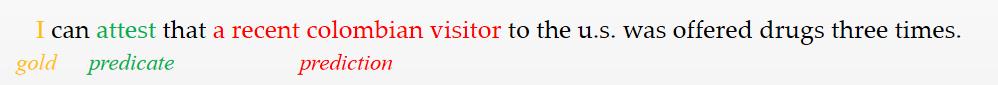}
        \caption{Illustration of a failure case for long dependency.}
        \label{fig:case_study_full_failure}
    \end{figure}

\subsubsection{Long-range dependency} 
    Long-range dependencies are always challenging for models. Even though \model works better compared with the baseline model, we indeed find failure cases, e.g., subject identification for main-subordinate complex sentences. 
    We visualize a failure case predicted by \model in Fig. \ref{fig:case_study_full_failure}. We also visualize a successful case in Fig. \ref{fig:case_study_full_success}. Look at the first sentence in Fig. \ref{fig:case_study_full_success}. The baseline model wrongly predicts ``justice marshall'' that should be ``associates of 81-year-old justice marshall'' in the ground truth. 
    The baseline model confused the position of the predicate in the subject-subordinate sentence and predict the wrong subject.
\subsubsection{Nuanced differences between similar arguments} LLMs are struggling to discern nuanced differences between similar arguments. For the following sentence \textit{``The Treasury plans to raise \$ 175 million in new cash Thursday by selling about \$ 9.75 billion of 52-week bills and redeeming \$ 9.58 billion of maturing bills.''}, \model wrongly identify the argument of \texttt{ADV} of the predicate ``raise'' as \textit{``by selling about \$ 9.75 billion of 52-week bills and redeeming \$ 9.58 billion of maturing bills''} which is actually the argument of \texttt{MNR}. The main reason for this failure is that it is difficult to distinguish the definition of \texttt{MNR} (\ie, the manner of the action) and \texttt{ADV} (\ie, the adverbial that modifies the predicate). The definition of \texttt{ADV} is complex. In some other instances, the manner of the sentence can act as the \texttt{ADV} role.
\subsubsection{Marginalized and abstract roles} Typical argument roles include \texttt{A0}, \texttt{A1}, as well as adjunctive roles indicating temporal, manner, \etc.
    As shown in Fig. \ref{fig:variants}, different argument roles have different levels of difficulty for LLMs. We can observe that for argument roles that are closely associated with the marked predicate, for example, \texttt{A0} and \texttt{A1}, the micro F1 of both can achieve surprisingly almost 60\% on ChatGPT. In contrast, when it comes to relatively marginalized and abstract roles, especially those rigorously defined by linguistic theory rather than intuitively described with simple natural language, the performance of \model decreases dramatically. For example, \model obtains only 12.21\% micro F1 for \texttt{DIR} on ChatGPT. 


\subsection{Impact of Exemplars}\label{sec:exp_exemplar}
\subsubsection{Compensation of examples to abstract roles}
Argument roles are abstract and symbolic abbreviations while the frame description is short and may lack information to explain the argument role to LLMs. 
We show the results of zero-/few-shot in Table \ref{tab:fewshot-result}. The results show that exemplars could bring tremendous performance improvement. 1-shot prompting outperforms 0-shot by a large margin of about 17\%.
\subsubsection{Trade-off between more exemplars and longer prompt}
From Table \ref{tab:fewshot-result}, 
ChatGPT achieves its highest F1 score when provided with a \textbf{3}-shot example. If we provide more than 3 exemplars, however, the performances do not increase significantly. We suspect that this is due to the long text input. 

\begin{table}
\vspace{-2em}
\small
\centering
\caption{Comparison of different shots for \model (\textbf{ChatGPT}) on CoNLL-2005 WSJ test set.}
\label{tab:fewshot-result}
\begin{tabular}{|c|ccc|} 
\hline
\multirow{2}{*}{\begin{tabular}[c]{@{}c@{}}\model (\textbf{ChatGPT})\end{tabular}} & \multicolumn{3}{c|}{\textbf{CoNLL-2005 WSJ}}  \\ 
\cline{2-4}
                         & P     & R     & F1                  \\ 
\hline
0-shot  & 27.08 & 18.71 & 22.13            \\
1-shot  & 39.42 & 38.85 & 39.13            \\
3-shot  & 39.19 & 41.73 & 40.42            \\
5-shot  & 38.56 & 42.45 & 40.41            \\
7-shot  & 38.78 & 41.01 & 39.86            \\
\hline
\end{tabular}
\vspace{-2em}
\end{table}

\begin{table}[!ht]
\small
\centering
\caption{Results of different prompts on ChatGPT.}
\label{tab:robust_result}
\begin{tabular}{|l|p{5.8cm}|ccc|ccc|} 
\hline
         & \multicolumn{1}{c|}{\multirow{2}{*}{Prompt template~}}                                                                                                                                                                                                                                                                                                                                                        & \multicolumn{3}{c|}{1-shot} & \multicolumn{3}{c|}{3-shot}  \\
         \cline{3-8}
 &     & P     & R     & F1        & P     & R     & F1         \\ 
\hline
Original & \begin{tabular}[c]{@{}p{5.8cm}@{}}Give a sentence: \textless{}sentence\textgreater{}, a predicate: \textless{}predicate\textgreater{}, all possible argument roles and their descriptions: \textless{}\{role: description\}\textgreater{}, please label the arguments about predicate \textless{}predicate\textgreater{} of the sentence. \\
Provide the answer in JSON format as follows: \{\textless{}predicate\textgreater{}: \{role: argument\}\}.\end{tabular}    & 39.42 & 38.85 & 39.13     & 39.19 & 41.73 & 40.42      \\ 
\hline
1        & \begin{tabular}[c]{@{}p{5.8cm}@{}}Give a sentence: \textless{}sentence\textgreater{}, all the possible arguments of
the prdicate <predicate> and their descriptions: \textless{}\{role: description\}\textgreater{}, please label the arguments about predicate \textless{}predicate\textgreater{} of the sentence. \\
Provide the answer in JSON format as follows: \{\textless{}predicate\textgreater{}: \{role: argument\}\}.\end{tabular} & 41.98 & 39.57 & 40.74     & 41.73 & 41.73 & 41.73      \\
\hline

\makecell[l]{2}       & \begin{tabular}[c]{@{}p{5.8cm}@{}}Give sentence: \textless{}sentence\textgreater{}, predicate: \textless{}predicate\textgreater{}, possible argument and descriptions:  \textless{}\{role: description\}\textgreater{}, please label the arguments about predicate \textless{}predicate\textgreater{} of the sentence. \\
Provide the answer in JSON format as follows: \{\textless{}predicate\textgreater{}: \{role: argument\}\}.\end{tabular}                                                                                                                                      & 35.66 & 33.09 & 34.33     & 39.1  & 37.41 & 38.24      \\
\hline
3        & \begin{tabular}[c]{@{}p{5.8cm}@{}}Give sentence: \textless{}sentence\textgreater{}, predicate: \textless{}predicate\textgreater{}, please label the arguments about predicate \textless{}predicate\textgreater{} of the sentence. \\
Provide the answer in JSON format as follows: \{\textless{}predicate\textgreater{}: \{role: argument\}\}.\end{tabular}                                                                                                                                                                                                                   & 31.91 & 32.37 & 32.14     & 29.66 & 30.94 & 30.28      \\
\hline
4        & \begin{tabular}[c]{@{}p{5.8cm}@{}}Give sentence: \textless{}sentence\textgreater{}, predicate: \textless{}predicate\textgreater{}, please label the arguments about predicate \textless{}predicate\textgreater{} of the sentence in the given PropBank-style semantic role labels. \\
Provide the answer in JSON format as follows: \{\textless{}predicate\textgreater{}: \{role: argument\}\}.\end{tabular}                                                                                                                & 41.94 & 37.41 & 39.54     & 46.04 & 46.04 & 46.04      \\
\hline
\end{tabular}
\vspace{-2.2em}
\end{table}

\subsection{Comparison With Untrained Humans}\label{sec:exp_compr_human_llm}
To investigate whether LLMs are human-like in making mistakes, we compare the LLM (ChatGPT, 3-shot) with untrained humans. All untrained humans (5 subjects) were senior students from universities who had passed the qualifications or evidence of English language ability (\eg, CET-6). We measure inter-annotator agreement using Fleiss' Kappa coefficient~\cite{fleiss1971measuring}, which is 0.71. Every subject was paid a wage of \$14.17/h. We released a questionnaire to collect untrained humans' SRL results. At the beginning of the questionnaire, there is a general description and several exemplars which is exactly the same as the prompt for LLMs. The subjects were required to write down their answers. We tested on 20 samples and manually compared the human results with LLM's predictions.
The F1 results of the LLM and untrained humans are 48.48\% and 70.10\%, respectively. Up to nearly 30\% of overlapping mistakes are made by both the LLM and untrained humans. Notely, the discussion on overlapping mistakes is based on \textbf{1-F1} score.
\subsubsection{Similarities}
ChatGPT and untrained humans both struggle to identify boundaries (error rate > 14.15\%). Additionally, they both incorrectly label certain arguments. For instance, in the sentence \textit{``Lily's grandmother, no cookie baker, excised the heads of disliked relatives from the family album, and lugged around her perennial work-in-progress, Philosophy for Women.''} with the predicate \textit{lugged}, the argument \textit{``around''} is labeled as location (\texttt{LOC}). Both of them failed.
\subsubsection{Differences}
Unlike untrained humans, ChatGPT is more likely to make mistakes in discontinuity. Given a sentence \textit{``The Bush administration earlier this year said it would extend steel quotas, known as voluntary restraint agreements, until March 31, 1992.''} with predicate \textit{``said''}, ChatGPT identifies the argument \textit{``The Bush administration earlier this year''} and labels it as \texttt{A0}. The arguments \textit{``The Bush administration''} and \textit{``earlier this year''} should be separately identified and respectively labeled as \texttt{A0} and \texttt{TMP}, as untrained humans do. Additionally, errors are often found in non-core arguments. We suspect that it is because they are not frequently present in corpora and, thus are more prone to be wrongly predicted by LLMs.

\subsection{Analysis of Different Prompts}\label{app:robust}

In this paper, we investigate the stability of the LLM results across different prompts, aiming to provide insights into the robustness of \model. Specifically, we conduct experiments on \model ChatGPT with various prompts and evaluate the performance on the CoNLL-2005 WSJ test dataset.

Four other different prompts are tested as shown in Table \ref{tab:robust_result}. Prompt 1 and Prompt 2 provide the same information compared to the original prompt, but differ in their presentation format. Prompt 1 provides a more direct and clear description of the relationship between the given arguments and the predicate, and Prompt 2 adopts a simpler syntax and omits the conditional statement. 
Prompt 3 and Prompt 4 adopt a similar presentation format as Prompt 2, and they do not include the same information compared to the original prompt. Prompt 3 omits the possible argument and description while Prompt 4 provides additional information ``PropBank-style''.

From Table \ref{tab:robust_result}, we find that: (i) A more direct and clear statement is beneficial even if two prompts are given the same information (Prompt 1 and 2); (ii) Prompts including more relevant information have a positive impact for ChatGPT (\eg, Prompt 3 and Prompt 4). To some extent, sufficient relevant information provides ChatGPT with additional contexts and cues to better understand and capture structured semantics.

\subsection{Ablation Study}
 We conduct the ablation study in Table \ref{tab:oblation-post-process} to illustrate the impact of the frame description. \model outperforms the counterpart without the frame description by around 4\% for micro F1. 

\begin{table}
    \vspace{-2em}
    \small
    \centering
    \caption{With \vs Without description in 3-shot setting for Davinci on CoNLL-2005 WSJ test set.}
    \label{tab:oblation-post-process}
    \begin{tabular}{|c|ccc|}
    \hline
         & \textbf{P} & \textbf{R} & \textbf{F1} \\ 
         \hline
    \model & 12.07 & 14.79 & 13.29   \\
    w/o description  &  5.65  & 25.35 & 9.25  \\ 
     \hline
    \end{tabular}
    
    \vspace{-2em}
    \end{table}
 
\section{Conclusion}
We first investigate LLMs' capability of extracting structured semantics by analyzing few-shot SRL. 
The structured semantics extraction reflects LLMs' expressive power.
We additionally assess their potential, limitations, and correlation to humans.
Additionally, our focus is analyzing and probing LLMs. This makes \model predict SRL arguments only depending on the inherent capability of LLMs instead of using external knowledge resources or tools except for a basic PropBank. Nevertheless, this makes it difficult to predict precise answers. It is difficult for \model to handle some complicated situations, like the C-argument and R-argument as well as some arguments with multiple meanings, because it is not easy to manually design the corresponding prompts for these situations.

%
%
%
\bibliographystyle{splncs04}
\bibliography{mybibfile,anthology}

\end{document}

%% file: config.tex
\usepackage{xspace}
\usepackage{url}

\usepackage[misc]{ifsym}

\usepackage{tabularx}
\usepackage{booktabs}
\usepackage{makecell}
\usepackage{multirow}
\usepackage{boldline}
\usepackage{caption}

\usepackage{wrapfig}
\usepackage{atbegshi}

\usepackage{algorithm}
\usepackage{algorithmic}

\usepackage{xcolor}

\newcommand{\model}{{\fontfamily{lmtt}\selectfont PromptSRL}\xspace}

\def\etc{{etc}}
\def\ie{{i.e.}}
\def\etal{{et al.}\xspace}
\def\vs{\emph{vs.}}
\def\eg{{e.g.}}